*Article*

# Inverse Airborne Optical Sectioning

**Rakesh John Amala Arokia Nathan[1], Indrajit Kurmi[2] and Oliver Bimber***

[1]  Johannes Kepler University Linz; rakesh_john.amala_arokia_nathan@jku.at
[2]  Johannes Kepler University Linz; indrajit.kurmi@jku.at
*  Johannes Kepler University Linz; Correspondence: oliver.bimber@jku.at; Tel.: +43-732-2468-6631

**Abstract:** We present Inverse Airborne Optical Sectioning (IAOS) – an optical analogy to Inverse Synthetic Aperture Radar (ISAR). Moving targets, such as walking people, that are heavily occluded by vegetation can be made visible and tracked with a stationary optical sensor (e.g., a hovering camera drone above forest). We introduce the principles of IAOS (i.e., inverse synthetic aperture imaging), explain how the signal of occluders can be further suppressed by filtering the Radon transform of the image integral, and present how targets' motion parameters can be estimated manually and automatically. Finally, we show that while tracking occluded targets in conventional aerial images is infeasible, it becomes efficiently possible in integral images that result from IAOS.

## 1. Introduction

Higher resolution, wide depth of field, fast framerates, high contrast or signal-to-noise ratio can often not be achieved with compact imaging systems that apply narrower aperture sensors. Synthetic aperture (SA) sensing is a widely recognized technique to achieve these objectives by acquiring individual signals of multiple or a single moving small-aperture sensor and by computationally combining them to approximate the signal of a physically infeasible, hypothetical wide aperture sensor [1]. This principle has been used in a wide range of applications, such as radar [2-28], telescopes [29,30], microscopes [31], sonar [32-35], ultrasound [36-37], lasers [38,39], and optical imaging [40-47].

In radar, electromagnetic waves are emitted and their backscattered echoes are recorded by an antenna. Electromagnetic waves at typical radar wavelengths (as compared to the visible spectrum) can penetrate scattering media (i.e., clouds, vegetation, and partly soil) and are quite useful for obtaining information in all weather conditions. However, acquiring high spatial resolution images would require an impractically large antenna [2]. Therefore, since its invention in 1950s [3,4], synthetic aperture radar (SAR) sensors have been placed on space-borne systems, like satellites [5-8], planes [9-11], and drones [12,13] in different modes of operation, such as strip-map [11,14], spotlight [11,14], and circular [10,14] to observe various sorts of phenomena on earth's surface. These include crop growth [8], mine detection [12], natural disasters [6], and climate change effects, such as the deforestation [14] or melting of glaciers [7]. Phase differences of multiple SAR recordings (interferometry) have even been used to reconstruct depth information and enables finer resolutions [15].

Analogous to SAR (which utilizes moving radars for synthetic aperture sensing of widely static targets), a technique known as inverse synthetic aperture radar (ISAR) [16-18] considers the relative motion of moving targets and static radars for SAR sensing. In contrast to SAR (where the radar motion is usually known), ISAR is challenged by the estimation of an unknown target motion. It often requires sophisticated signal processing and is often limited to sensing one target at a time, while SAR can image large areas and monitor multiple (static) targets simultaneously [17,18]. ISAR has been used for non-co-operative target recognition (non-stationary targets) in maritime [19,20], airspace [21,22], near-space [23,24], and overland surveillance applications [25-28]. Recently, spatially distributed systems and advanced signal processing, such as compressed sensing and machine learning, have been utilized to obtain 3-D images of targets, target's reflectivity, and more degrees of freedom for target motion estimation [27,28].

With Airborne Optical Sectioning (AOS) [48-60], we have introduced an optical synthetic aperture imaging technique that captures an unstructured light field with an aircraft, such as a drone. So far, we utilized manually, automatically [48-55,57-59], or fully autonomously [56] operated camera drones that sample multispectral (RGB and thermal) images within a certain (synthetic aperture) area above occluding vegetation (such as forest) and combine their signals computationally to remove occlusion. The outcome is a widely occlusion-free integral image of the ground, revealing details of registered targets while unregistered occluders above the ground, such as trunks, branches or leaves disappear in strong defocus. In contrast to SAR, AOS benefits from high spatial resolution, real-time processing rates, and wavelength-independences – making it useful in many domains. So far, AOS has been applied to the visible [48,59] and the far-infrared (thermal) spectrum [51] for various applications, such as archeology [48,49], wildlife observation [52], and search and rescue [55,56]. By employing a randomly distributed statistical model [50,57,60] the limits of AOS and its efficacy with respect to its optimal sampling parameters can be explained. Common image processing tasks, such as classification with deep neural networks [55,56] or color anomaly detection [59] have been proven to perform significantly better when applied to AOS integral images as compared to conventional aerial images. We have also demonstrated the real-time capability of AOS by deploying it on a fully autonomous and classification-driven adaptive search and rescue drone [56]. Yet, the sequential sampling nature of AOS when being used with conventional single-camera drones has limited its applications to recover static targets only. Moving targets lead to motion blur in the AOS integral images, which is nearly impossible to classify or to track.

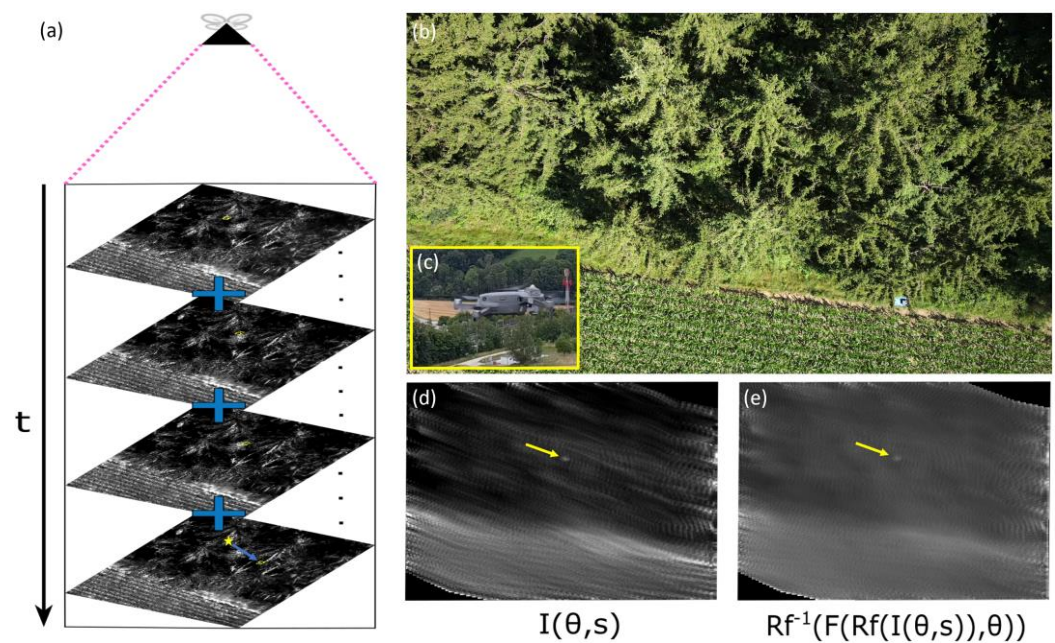

$$I(\theta,s) \qquad Rf^{-1}(F(Rf(I(\theta,s)),\theta))$$

**Figure 1.** Inverse Airborne Optical Sectioning (IAOS) principle: IAOS relies on the motion of targets being sensed by a static airborne optical sensor (e.g., a drone (c) hovering above forest (b)) over time (a) to computationally reconstruct an occlusion free integral image I (d). Essential for an efficient reconstruction is the correct estimation of the target's motion (direction $\theta$, and speed s). By filtering the Radon transform of I, the signal of occluders can be suppressed further (e). Thermal images are shown in (a,d,e).

In [59], we presented a first solution to tracking moving people through densely occluding foliage with parallel synthetic aperture sampling supported by a drone-operated, 10m wide, 1D camera array (assembling 10 synchronized cameras). Although proofing feasibility, such a specialized imaging system is in most use-cases impractical, as it is bulky and difficult to control.

Being inspired by the principles of ISAR for radar, we present Inverse Airborne Optical Sectioning (IAOS) for detecting and tracking moving targets through occluding foliage (cf. Fig. 1b) with a conventional, single-camera drone (cf. Fig. 1c). As with ISAR, IAOS relies on the motion of targets being sensed by a static airborne optical sensor (e.g., a drone hovering above forest) over time (cf. Fig. 1a) to computationally reconstruct an occlusion free integral image (cf. Fig. 1d). Essential for an efficient reconstruction is the correct estimation of the target's motion.

In this article, we make four main contributions: (1) We introduce the principles of IAOS (i.e., inverse synthetic aperture imaging). (2) We explain how the signal of occluders can be further suppressed by filtering the Radon transform of the image integral (cf. Fig. 1e). (3) We present how targets' motion parameters can be estimated manually and automatically. (4) We show that while tracking occluded targets in conventional aerial images is infeasible, it is efficiently possible in integral images that result from IAOS.

## 2. Materials and Methods

All field experiments were carried out in compliance with the legal European union Aviation Safety Agency (EASA) flight regulations, using a DJI Mavic 2 Enterprise Advanced, over dense broad-leaf, conifer, and mixed forest, and under direct sunlight as well as under cloudy weather conditions. Free flight drone operations were performed using the DJI's standalone smart remote controller with DJI's Pilot application. RGB videos of resolution 1920 x 1080 (30 fps) and thermal videos of resolution 640 x 512 (30 fps) were recorded on the drone's internal memory, and were processed offline after landing. For vertical (top-down, as in Fig. 3) scans the drone was hovering at an altitude of about 35m AGL. For horizontal scans (sideways, as in Fig. 4) the drone was hovering at a distance of about 10m away from the vegetation. For quicker processing, we extracted a selection of 1-5 fps from the acquired 30 fps thermal videos using FFmpeg python bindings. Offline processing included intrinsic camera calibration (pre-calibrated transformation matrix computed using MATLAB's camera calibrator application) and image un-distortion/rectification using OpenCV's pinhole camera model (as explained in [48,55]). The undistorted and rectified images were cropped to a field of view of 36° and a resolution of 1024px × 1024px. Image integration was achieved by averaging the pre-processed images being registered based on manually or automatically estimated motion parameters, as explained in sections 2.1 and 2.2. Radon transform filtering [61-63] (also explained in sections 2.1 and 2.2) was implemented in MATLAB.

### 2.1. Manual Motion Estimation

If the target's motion parameters (i.e., direction, θ [°] and speed, s [m/s]) are known and assumed to be constant for all time intervals, the captured images can be registered by shifting them accordingly to θ and s. Thereby, θ can directly be mapped to the image plane, while s has to be mapped [m/s] to [px/s] (which is easily possible after camera calibration and knowing the drone's altitude). By averaging the registered images results in an integral image that shows the target in focus (note, that local motion of the target itself, such as arm movements of a walking person, still lead to defocus) while the misregistered occluders vanish in defocus.

Large occluders that are shifted in direction θ while being integrated appear as linear directional blur artifacts in the integral image (cf. Fig. 1d). Their signal can be suppressed by filtering (zeroing out) the Radon transform of the integral image $I(\theta,s)$ in direction θ (+/- an uncertainty range that considers local motion non-linearities of the occluders, such as movements of branches caused by wind, etc.). The inverse Radon transform (filtered back projection [63]) of the filtered sinogram results in a new integral image with suppressed signal of the directionally blurred occluders (cf. Fig. 1e). This process is illustrated in figure 2, and can be summarized mathematically with

$$I'(\theta, s) = Rf^{-1}\Big(F\big(Rf(I(\theta, s)), \theta\big)\Big), \tag{1}$$

where F is the filter function which zeros out coefficients at angle $\theta$ (+/- uncertainty range) in the sinogram.

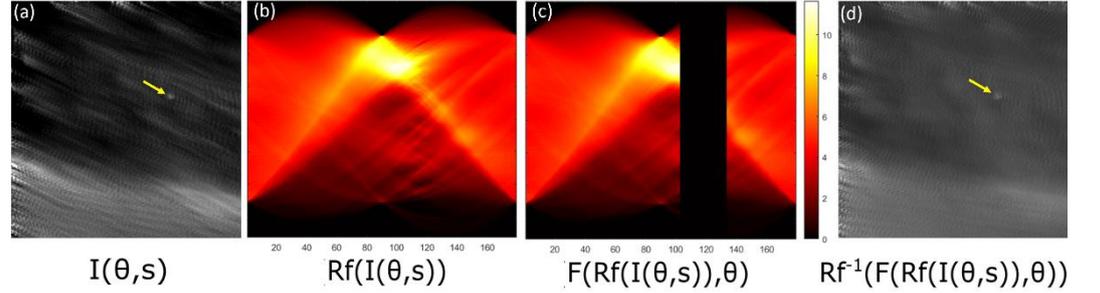

$$I(\theta,s) \qquad Rf(I(\theta,s)) \qquad F(Rf(I(\theta,s)),\theta) \qquad Rf^{-1}(F(Rf(I(\theta,s)),\theta))$$

**Figure 2.** Radon transform filtering: To suppress directional blur artifacts of large occluders integrated in direction $\theta$ (a), the Radon transform (Rf) of the integral image (b) is filtered with function F that zeros out $\theta$, +/- an uncertainty range which takes local motion of the occluders themselves into account (c). The inverse Radon transform ($Rf^{-1}$) of this filtered sinogram suppresses the direction blur artefacts of the occluders (d). Note, that remaining directional artifacts in orthogonal directions are caused by under-sampling (i.e., the number of images being integrated). They are fluctuating too much to be suppressed in the same manner. In the example above, $\theta$=118° with +/- 15° (image coordinate system: clockwise, +y-axis = 0°).

One way of estimating the correct motion parameters is by visual search (i.e., $\theta$ and s are interactively modified until the target appears best focused in the integral image). Exploring the two-dimensional parameter space within proper bounds is relatively efficient if the motion can be assumed to be constant. Sample results are presented in section 3. See also supplementary Video 1 for an example of manual visual search for the motion parameters of results shown in Fig. 3k. In case of non-linear motion, the motion parameters have to be continuously and automatically estimated. A manual exploration becomes infeasible in this case.

### 2.2. Automatic Motion Estimation

Automatic estimation of motion parameters requires an error metric which is capable of detecting improvement and degradation in visibility (i.e., focus and occlusion) for different parameters. Here, we utilize simple gray level variance (GLV) [64] as an objective function. We have already proven in [53] that, in contrast to traditionally used gradient-, Laplacian-, or wavelet-based focus metrics [65], GLV does not rely on any image features and is thus invariant to occlusion. In [54] (see also Appendix), we demonstrated that the variance of an integral image is

$$Var[I] = \frac{D(1-D)((\mu_o - \mu_s)^2) + D\sigma_o^2 + (1-D)\sigma_s^2}{N} + (1-D)^2\Big(1 - \frac{1}{N}\Big)\sigma_s^2, \tag{2}$$

where D is the probability of occlusion, while $\mu_o$, $\sigma_o^2$ and $\mu_s$, $\sigma_s^2$ are the statistical properties of occlusion and the target signal respectively.

Integrating N individual images with optimal motion parameters will result in occlusion-free view of the target's signal whereas the signal strength of the occluders will reduce and disappear in strong defocus. To further suppress occluders, we use Radon filtering [61-63] as described in section 2.1. However, we now utilize the linearity property of the Radon transform which states that

$$Rf(\sum_i \alpha_i I_i) = \sum_i \alpha_i Rf(I_i). \qquad (3)$$

Thus, instead of filtering the integral image $I(\theta,s)$, as explained in Eqn. 1, we apply Radon transform filtering to each single image $I_i$ before integrating it.

For automatic motion parameter estimation, we register the current integral image I (integrating $I'_1 \dots I'_{i-1}$) to the latest (most recently recorded) inverse Radon transformed filtered image $I'_i = Rf^{-1}(F(Rf(I_i), \theta))$ by maximizing Var[I] while optimizing for best motion parameters $(\theta, s)$. Deterministic-global search, DIRECT [66] (as implemented Nlopt [67]), is applied for optimization. Consequently, we consider each discrete motion component between two recorded images and within the corresponding imaging time (e.g., 1/30s for 30fps) to be piecewise linear. The integration of multiple images, however, can reveal and a track a non-linear motion pattern where $(\theta, s)$ vary in each recording step. Sample results are presented in section 3 and in supplementary Videos 2 and 3.

## 3. Results

Figure 3 presents results from field studies of IAOS with manual motion estimation, as explained in section 2.1. Images have been recorded top-down, with the drone hovering at a constant position above conifer (a-l), broadleaf (m-o), and mixed (p-r) forest. Estimated motion parameters of hidden walking people were: 118°,0.5m/s (Figs. 3a-l), 108°,0.6m/s (Figs. 3m-o), and 90°, 0.6m/s (Figs. 3p-q).

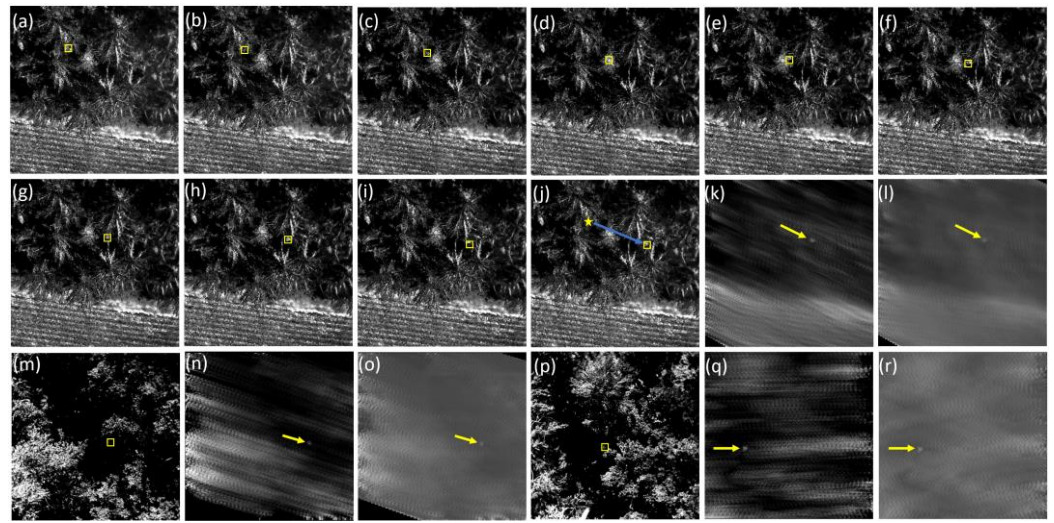

**Figure 3.** Manual motion estimation (vertical): Sequence of single thermal images (a-j) with walking persons indicated (yellow box), distance covered by person during capturing time (j), computed integral image (k), and Radon transform filtered integral image (l). Target indicated by yellow arrow. Different forest types: single thermal image example (m,p), integral images (n,q), and Radon transform filtered integral images (o,r).

Figure 4 illustrates an example with the drone hovering at a distance of 10m in front of dense bushes (at an altitude of 2m, recording horizontally). The hidden person is walking from right to left at 260° with 0.27m/s (both manually estimated).

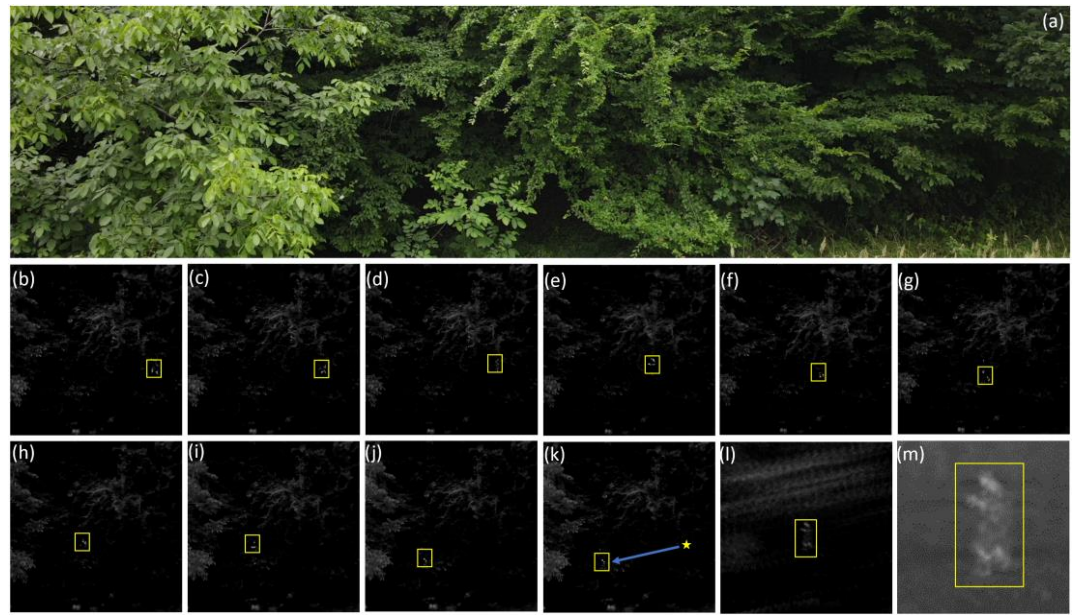

**Figure 4.** Manual motion estimation (horizontal): Walking person behind dense bushes. RGB image of drone (a). Single thermal images with person position indicated with yellow box (b-k). Distance covered by person during capturing time (k). Integral image (l) and close up (m) where the shape of the person can be recognized.

Figure 5 illustrates two examples for automatic motion estimation, as explained in section 2.2, with the drone hovering at an altitude of 35m and a hidden person walking through dense forest.

For tracking, moving targets are first detected by utilizing background subtraction based on Gaussian mixture models [68,69]. The resulting foreground mask is further processed using morphological operations to eliminate noise [70,71]. Subsequently blob analysis [72,73] detects connected pixels corresponding to each moving target. Association of detections in subsequent frames is entirely based on motion where the motion of each detected target is estimated by a Kalman filter. The filter predicts the target location in subsequent frame (based on previous motion and associated motion model) and then determines the likelihood of assigning the detection to the target.

For comparison, we apply the above tracking approach to both: the sequence of captured single thermal images and to the sequence of integral images computed from the single images, as described in section 2.2. For each case, tracking parameters (such as minimum blob size, max. prediction length, no. training images for background subtraction) are individually optimized to achieve best possible results.

While tracking in single images leads to many false positive detections and becomes practically infeasible, tracking in integral images results in clear track-paths of a single target. Estimated mean motion parameters were: 291°,0.82m/s (Figs. 5a-c), 309°,0.16m/s for the first leg and 241°,0.41m/s for the second leg (Figs. 5d-f). See supplementary Videos 2 and 3 for dynamic examples of these results.

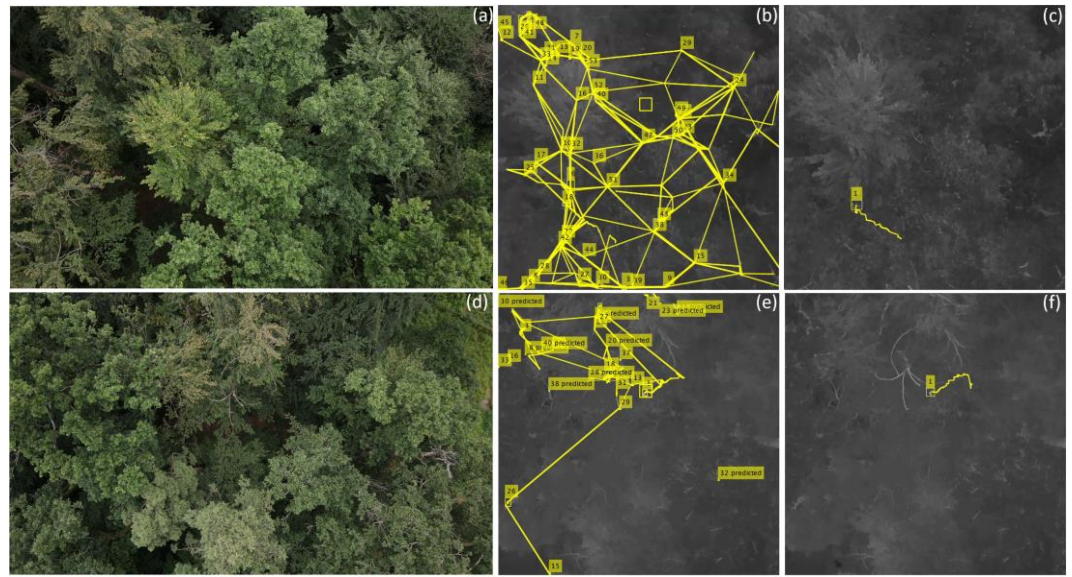

**Figure 5.** Automatic motion estimation (vertical): Two examples of tracking a moving hidden person within dense forest (a,d: RGB images of drone) in either single thermal images (b,e) and IAOS integral images. Note, that the tracking results of the integral images were projected back to a single thermal image for better spatial reference (c,f). Motion paths are indicated by yellow lines. While tracking in single images leads to many false positive detections, tracking in integral images results in clear track-paths of a single target. See supplementary Videos 2 and 3 for dynamic examples of these results.

## 4. Discussion and Conclusion

In this article we presented Inverse Airborne Optical Sectioning (IAOS) – an optical analogy to Inverse Synthetic Aperture Radar (ISAR). Moving targets, such as walking people, that are heavily occluded by vegetation can be made visible and tracked with a stationary optical sensor (e.g., a hovering camera drone above forest). We introduced the principles of IAOS (i.e., inverse synthetic aperture imaging), explained how the signal of occluders can be further suppressed by filtering the Radon transform of the image integral, and presented how targets' motion parameters can be estimated manually and automatically. Furthermore, we showed that while tracking occluded targets in conventional aerial images is infeasible, it is efficiently possible in integral images that result from IAOS.

IAOS has several limitations: We assume, that local motion of occluders and of the drone (e.g., caused by wind) is smaller than the motion of the target. Small local motion of the target itself, such as individual moving body parts, still appear blurred in integral images. The field of view of a hovering drone is limited, and moving targets might be out of view quickly. In future, we will investigate how drone movement being adapted to target movement can increase field of view and reduce blur of local target motion. This corresponds to a combination of IAOS (i.e., occlusion removal by registering target motion) and classical AOS (i.e., occlusion removal by registering drone movement). Furthermore, results of Radon transform filtering have artifacts that are due to under-sampling. Higher imaging rates will overcome this. The blob-based tracking approach applied for proof-of-concept is very simple. More sophisticated methods will achieve superior See supplementary Videos 2 and 3 for dynamic examples of these results tracking-results. However, we believe that tracking in integral images will always outperform tracking in conventional images.

**Supplementary Materials:** The following supporting information can be downloaded at: https://github.com/JKU-ICG/AOS/ , Video 1: Manual visual search for the motion parameters. Video 2: Automatic motion estimation (example 1). Video 3: Automatic motion estimation (example 2).

**Author Contributions:** Conceptualization, O.B.; methodology, O.B., R.N., and I.K.; software, R.N., I.K..; validation, R.N., I.K. and O.B.; formal analysis, R.N., I.K. and O.B.; investigation, R.N. and I.K.; resources, R.N. and I.K.; data curation, R.N. and I.K.; writing—original draft preparation, O.B., R.N., and I.K.; writing—review and editing,, O.B., R.N., and I.K.; visualization, O.B., R.N., and I.K.; supervision, O.B.; project administration, O.B.; funding acquisition, O.B. All authors have read and agreed to the published version of the manuscript.

**Funding:** This research was funded by the Austrian Science Fund (FWF) under grant number P 32185-NBL, and by the State of Upper Austria and the Austrian Federal Ministry of Education, Science and Research via the LIT–Linz Institute of Technology under grant number LIT-2019-8-SEE-114.

**Data Availability Statement:** The data, code, and supplementary material are available on GitHub: https://github.com/JKU-ICG/AOS/

**Acknowledgments:** We want to thank the Upper Austrian Fire Brigade Headquarters for providing the DJI Mavic 2 Enterprise Advanced for our experiments.

**Conflicts of Interest:** The authors declare no conflict of interest.

**Appendix**

In the following, we present the derivation of an integral image's variance ($Var[I]$). We apply the statistical model described in [50], where the integral image $I$ is composed of $N$ single image recordings $I_i$ and each single image pixel in $I_i$ is either occlusion free ($S$) or occluded ($O$), determined by $Z$:

$$I_i = Z_i O_i + (1 - Z_i)S.$$

Similar as in [50], all variables are independent and identically distributed with $Z_i$, following a Bernoulli distribution with success parameter $D$ (i.e., $E[Z_i] = E[Z_i^2] = D$; furthermore note, that $E[Z_i(1 - Z_i)] = 0$ is true). The random variable S follows a distribution whose properties can be described with mean $E[S] = \mu_s$ and $E[S^2] = (\mu_s^2 + \sigma_s^2)$. Analogously, the occluded variable $O_i$ follows a distribution whose properties can be described with $E[O_i] = \mu_o$ and $E[O_i^2] = (\mu_o^2 + \sigma_o^2)$. We compute the first and second moments of $I_i$ to determine its mean and variance with

$$E[I_i] = D\mu_o + (1 - D)\mu_s$$

and

$$E[I_i^2] = D(\mu_o^2 + \sigma_o^2) + (1 - D)(\mu_s^2 + \sigma_s^2).$$

Variances of single images $I_i$ can be obtained as

$$\begin{aligned}
Var[I_i] &= E[I_i^2] - (E[I_i])^2 \\
&= D(\mu_o^2 + \sigma_o^2) + (1 - D)(\mu_s^2 + \sigma_s^2) \\
&\quad - (D^2\mu_o^2 + (1 - D)^2\mu_s^2 + 2D(1 - D)\mu_o\mu_s \\
&= D(1 - D)((\mu_o - \mu_s)^2) + D\sigma_o^2 + (1 - D)\,\sigma_s^2 \ .
\end{aligned}$$

Similarly, for I, we determine the first and second moments where the first moment of I is given by

$$\begin{aligned}
E[I] &= E\left[\frac{1}{N}\sum_{i=1}^{N} Z_i O_i + (1 - Z_i)S\right] \\
&= D\mu_o + (1 - D)\mu_s
\end{aligned}$$

and the second moment of I is as derived in [50]

$$E[I^2] = \frac{1}{N^2}\left(\begin{array}{l} N\big(D(\sigma_o^2 + \mu_o^2) + (1 - D)(\sigma_s^2 + \mu_s^2)\big) \\ + N(N - 1)\begin{pmatrix} D^2\mu_o^2 + 2D(1 - D)\mu_o\mu_s \\ + (1 - D)^2(\sigma_s^2 + \mu_s^2) \end{pmatrix} \end{array}\right).$$

Consecutively, we calculate the variance of the integral image as

$$\begin{aligned}
Var[I] &= E[I^2] - (E[I])^2 \\
&= \frac{1}{N}\big(D(\sigma_o^2 + \mu_o^2) + (1 - D)(\sigma_s^2 + \mu_s^2)\big) \\
&\quad + \big(D^2\mu_o^2 + 2D(1 - D)\mu_o\mu_s + (1 - D)^2(\sigma_s^2 + \mu_s^2)\big) \\
&\quad - \frac{1}{N}\big(D^2\mu_o^2 + 2D(1 - D)\mu_o\mu_s + (1 - D)^2(\sigma_s^2 + \mu_s^2)\big) \\
&\quad - \big(D^2\mu_o^2 + (1 - D)^2\mu_s^2 + 2D(1 - D)\mu_o\mu_s\big) \\
&= \frac{1}{N}\big(D(\sigma_o^2 + \mu_o^2) + (1 - D)(\sigma_s^2 + \mu_s^2)\big) + (1 - D)^2\sigma_s^2 \\
&\quad - \frac{1}{N}\big(D^2\mu_o^2 + 2D(1 - D)\mu_o\mu_s + (1 - D)^2(\sigma_s^2 + \mu_s^2)\big)
\end{aligned}$$

$$= \frac{1}{N}(D(1-D)((\mu_o - \mu_s)^2) + D\sigma_o^2 + (1-D)\sigma_s^2)$$
$$+ (1-D)^2(1 - \frac{1}{N})\sigma_s^2 \ .$$

## References

1. Ryle, M., and D. D. Vonberg. "Solar radiation on 175 Mc./s." *Nature* . **158**, 1946. 339–340.
2. Moreira, Alberto, et al. "A tutorial on synthetic aperture radar." *IEEE Geoscience and remote sensing magazine* 1.1 (2013): 6-43.
3. Wiley, Carl A. "Pulsed doppler radar methods and apparatus." U.S. Patent No. 3,196,436. 20 Jul. 1965.
4. Cutrona, L., et al. "Synthetic aperture radars: A paradigm for technology evolution." *IRE Trans. Military Electron* (1961): 127-131.
5. Farquharson, Gordon, et al. "The capella synthetic aperture radar constellation." *EUSAR 2018; 12th European Conference on Synthetic Aperture Radar*. VDE, 2018.
6. Chen, Fulong, Rosa Lasaponara, and Nicola Masini. "An overview of satellite synthetic aperture radar remote sensing in archaeology: From site detection to monitoring." *Journal of Cultural Heritage* 23 (2017): 5-11.
7. Zhang, Zhengjia, et al. "A Review of Satellite Synthetic Aperture Radar Interferometry Applications in Permafrost Regions: Current Status, Challenges, and Trends." *IEEE Geoscience and Remote Sensing Magazine* (2022).
8. Ranjan, Avinash Kumar, and Bikash Ranjan Parida. "Predicting paddy yield at spatial scale using optical and Synthetic Aperture Radar (SAR) based satellite data in conjunction with field-based Crop Cutting Experiment (CCE) data." *International journal of remote sensing* 42.6 (2021): 2046-2071.
9. Reigber, Andreas, et al. "Very-high-resolution airborne synthetic aperture radar imaging: Signal processing and applications." *Proceedings of the IEEE* 101.3 (2012): 759-783.
10. Sumantyo, Josaphat Tetuko Sri, et al. "Airborne circularly polarized synthetic aperture radar." *IEEE Journal of Selected Topics in Applied Earth Observations and Remote Sensing* 14 (2020): 1676-1692.
11. Tsunoda, Stanley I., et al. "Lynx: A high-resolution synthetic aperture radar." *2000 IEEE Aerospace Conference. Proceedings (Cat. No. 00TH8484)*. Vol. 5. IEEE, 2000.
12. Fernández, María García, et al. "Synthetic aperture radar imaging system for landmine detection using a ground penetrating radar on board a unmanned aerial vehicle." *IEEE Access* 6 (2018): 45100-45112.
13. Deguchi, Tomonori, Tomoyuki Sugiyama, and Munemaru Kishimoto. "Development of SAR system installable on a drone." *EUSAR 2021; 13th European Conference on Synthetic Aperture Radar*. VDE, 2021.
14. Mondini, Alessandro Cesare, et al. "Landslide failures detection and mapping using Synthetic Aperture Radar: Past, present and future." *Earth-Science Reviews* 216 (2021): 103574..
15. Rosen, Paul A., et al. "Synthetic aperture radar interferometry." *Proceedings of the IEEE* 88.3 (2000): 333-382.
16. Prickett, M. J., and C. C. Chen. "Principles of inverse synthetic aperture radar/ISAR/imaging." *EASCON'80; Electronics and Aerospace Systems Conference*. 1980.
17. Vehmas, Risto, and Nadav Neuberger. "Inverse Synthetic Aperture Radar Imaging: A Historical Perspective and State-of-the-Art Survey." *IEEE Access* (2021).
18. Ozdemir, Caner. *Inverse synthetic aperture radar imaging with MATLAB algorithms*. Vol. 210. John Wiley & Sons, 2012.
19. Marino, Armando, et al. "Ship detection with spectral analysis of synthetic aperture radar: A comparison of new and well-known algorithms." *Remote Sensing* 7.5 (2015): 5416-5439.
20. Wang, Yong, and Xuefei Chen. "3-D interferometric inverse synthetic aperture radar imaging of ship target with complex motion." *IEEE Transactions on Geoscience and Remote Sensing* 56.7 (2018): 3693-3708.
21. Xu, Gang, et al. "Sparse Inverse Synthetic Aperture Radar Imaging Using Structured Low-Rank Method." *IEEE Transactions on Geoscience and Remote Sensing* 60 (2021): 1-12.
22. Berizzi, Fabrizio, and Giovanni Corsini. "Autofocusing of inverse synthetic aperture radar images using contrast optimization." *IEEE Transactions on Aerospace and Electronic Systems* 32.3 (1996): 1185-1191.
23. Bai, Xueru, et al. "Scaling the 3-D image of spinning space debris via bistatic inverse synthetic aperture radar." *IEEE Geoscience and Remote Sensing Letters* 7.3 (2010): 430-434.
24. Anger, Simon, et al. "Research on advanced space surveillance using the IoSiS radar system." *EUSAR 2021; 13th European Conference on Synthetic Aperture Radar*. VDE, 2021.
25. Vossiek, Martin, et al. "Inverse synthetic aperture secondary radar concept for precise wireless positioning." *IEEE Transactions on Microwave Theory and Techniques* 55.11 (2007): 2447-2453.
26. Jeng, Shyr-Long, Wei-Hua Chieng, and Hsiang-Pin Lu. "Estimating speed using a side-looking single-radar vehicle detector." *IEEE transactions on intelligent transportation systems* 15.2 (2013): 607-614.
27. Ye, Xingwei, et al. "Photonics-based high-resolution 3D inverse synthetic aperture radar imaging." *IEEE Access* 7 (2019): 79503-79509.
28. Pandey, Neeraj, and Shobha Sundar Ram. "Classification of automotive targets using inverse synthetic aperture radar images." *IEEE Transactions on Intelligent Vehicles* (2022).
29. Levanda, Ronny, and Amir Leshem. "Synthetic aperture radio telescopes." *IEEE Signal Processing Magazine* 27.1 (2009): 14-29.

30. Dravins, Dainis, Tiphaine Lagadec, and Paul D. Nuñez. "Optical aperture synthesis with electronically connected telescopes." *Nature communications* 6.1 (2015): 1-5.
31. Ralston, Tyler S., et al. "Interferometric synthetic aperture microscopy." *Nature physics* 3.2 (2007): 129-134.
32. Edgar, Roy. "Introduction to Synthetic Aperture Sonar." *Sonar Systems* (2011): 1-11.
33. Hayes, Michael P., and Peter T. Gough. "Synthetic aperture sonar: A review of current status." *IEEE journal of oceanic engineering* 34.3 (2009): 207-224.
34. Hansen, Roy Edgar, et al. "Challenges in seafloor imaging and mapping with synthetic aperture sonar." *IEEE Transactions on geoscience and Remote Sensing* 49.10 (2011): 3677-3687.
35. Bülow, Heiko, and Andreas Birk. "Synthetic aperture sonar (SAS) without navigation: Scan registration as basis for near field synthetic imaging in 2D." *Sensors* 20.16 (2020): 4440.
36. Jensen, Jørgen Arendt, et al. "Synthetic aperture ultrasound imaging." *Ultrasonics* 44 (2006): e5-e15.
37. Zhang, Haichong K., et al. "Synthetic tracked aperture ultrasound imaging: design, simulation, and experimental evaluation." *Journal of Medical Imaging* 3.2 (2016): 027001.
38. Barber, Zeb W., and Jason R. Dahl. "Synthetic aperture ladar imaging demonstrations and information at very low return levels." *Applied optics* 53.24 (2014): 5531-5537.
39. Turbide, Simon, et al. "Synthetic aperture lidar as a future tool for earth observation." *International Conference on Space Optics—ICSO 2014*. Vol. 10563. SPIE, 2017.
40. Vaish, Vaibhav, et al. "Using plane+ parallax for calibrating dense camera arrays." *Proceedings of the 2004 IEEE Computer Society Conference on Computer Vision and Pattern Recognition, 2004. CVPR 2004.*. Vol. 1. IEEE, 2004.
41. Vaish, Vaibhav, et al. "Reconstructing occluded surfaces using synthetic apertures: Stereo, focus and robust measures." *2006 IEEE Computer Society Conference on Computer Vision and Pattern Recognition (CVPR'06)*. Vol. 2. IEEE, 2006.
42. Zhang, Heng, Xin Jin, and Qionghai Dai. "Synthetic aperture based on plenoptic camera for seeing through occlusions." *Pacific Rim Conference on Multimedia*. Springer, Cham, 2018.
43. Yang, Tao, et al. "Kinect based real-time synthetic aperture imaging through occlusion." *Multimedia Tools and Applications* 75.12 (2016): 6925-6943.
44. Joshi, Neel, et al. "Synthetic aperture tracking: tracking through occlusions." *2007 IEEE 11th International Conference on Computer Vision*. IEEE, 2007.
45. Pei, Zhao, et al. "Occluded-object 3D reconstruction using camera array synthetic aperture imaging." *Sensors* 19.3 (2019): 607.
46. Yang, Tao, et al. "All-in-focus synthetic aperture imaging." *European Conference on Computer Vision*. Springer, Cham, 2014.
47. Pei, Zhao, et al. "Synthetic aperture imaging using pixel labeling via energy minimization." *Pattern Recognition* 46.1 (2013): 174-187.
48. Kurmi, Indrajit, David C. Schedl, and Oliver Bimber. "Airborne optical sectioning." *Journal of Imaging* 4.8 (2018): 102.
49. Bimber, Oliver, Indrajit Kurmi, and David C. Schedl. "Synthetic aperture imaging with drones." *IEEE computer graphics and applications* 39.3 (2019): 8-15.
50. Kurmi, Indrajit, David C. Schedl, and Oliver Bimber. "A statistical view on synthetic aperture imaging for occlusion removal." *IEEE Sensors Journal* 19.20 (2019): 9374-9383.
51. Kurmi, Indrajit, David C. Schedl, and Oliver Bimber. "Thermal airborne optical sectioning." *Remote Sensing* 11.14 (2019): 1668.
52. Schedl, David C., Indrajit Kurmi, and Oliver Bimber. "Airborne optical sectioning for nesting observation." *Scientific reports* 10.1 (2020): 1-7.
53. Kurmi, Indrajit, David C. Schedl, and Oliver Bimber. "Fast automatic visibility optimization for thermal synthetic aperture visualization." *IEEE Geoscience and Remote Sensing Letters* 18.5 (2020): 836-840.
54. Kurmi, Indrajit, David C. Schedl, and Oliver Bimber. "Pose error reduction for focus enhancement in thermal synthetic aperture visualization." *IEEE Geoscience and Remote Sensing Letters* 19 (2021): 1-5.
55. Schedl, David C., Indrajit Kurmi, and Oliver Bimber. "Search and rescue with airborne optical sectioning." *Nature Machine Intelligence* 2.12 (2020): 783-790.
56. Schedl, David C., Indrajit Kurmi, and Oliver Bimber. "An autonomous drone for search and rescue in forests using airborne optical sectioning." *Science Robotics* 6.55 (2021): eabg1188.
57. Kurmi, Indrajit, David C. Schedl, and Oliver Bimber. "Combined person classification with airborne optical sectioning." *Scientific reports* 12.1 (2022): 1-11.
58. Ortner, Rudolf, Indrajit Kurmi, and Oliver Bimber. "Acceleration-Aware Path Planning with Waypoints." *Drones* 5.4 (2021): 143.
59. Amala Arokia Nathan, Rakesh John, et al. "Through-Foliage Tracking with Airborne Optical Sectioning." *Journal of Remote Sensing* 2022, https://doi.org/10.34133/2022/9812765 (2022).
60. Seits, Francis, et al. "On the Role of Field of View for Occlusion Removal with Airborne Optical Sectioning." *arXiv preprint arXiv:2204.13371* (2022).
61. Bracewell, Ronald N. *Two-dimensional imaging*. Prentice-Hall, Inc., 1995.
62. Lim, Jae S. "Two-dimensional signal and image processing." *Englewood Cliffs* (1990).
63. Kak, Avinash C., and Malcolm Slaney. *Principles of computerized tomographic imaging*. Society for Industrial and Applied Mathematics, 2001.
64. Firestone, Lawrence, et al. "Comparison of autofocus methods for automated microscopy." *Cytometry: The Journal of the International Society for Analytical Cytology* 12.3 (1991): 195-206.

65. Pertuz, Said, Domenec Puig, and Miguel Angel Garcia. "Analysis of focus measure operators for shape-from-focus." *Pattern Recognition* 46.5 (2013): 1415-1432.

66. Jones, Donald R., Cary D. Perttunen, and Bruce E. Stuckman. "Lipschitzian optimization without the Lipschitz constant." *Journal of optimization Theory and Applications* 79.1 (1993): 157-181.

67. Steven G. Johnson, The NLopt nonlinear-optimization package, http://github.com/stevengj/nlopt

68. KaewTraKulPong, Pakorn, and Richard Bowden. "An improved adaptive background mixture model for real-time tracking with shadow detection." *Video-based surveillance systems*. Springer, Boston, MA, 2002. 135-144.

69. Stauffer, Chris, and W. Eric L. Grimson. "Adaptive background mixture models for real-time tracking." *Proceedings. 1999 IEEE computer society conference on computer vision and pattern recognition (Cat. No PR00149)*. Vol. 2. IEEE, 1999.

70. Soille, Pierre. *Morphological image analysis: principles and applications*. Vol. 2. No. 3. Berlin: Springer, 1999.

71. Dougherty, Edward R., and Roberto A. Lotufo. *Hands-on morphological image processing*. Vol. 59. SPIE press, 2003.

72. Dillencourt, Michael B., Hanan Samet, and Markku Tamminen. "A general approach to connected-component labeling for arbitrary image representations." *Journal of the ACM (JACM)* 39.2 (1992): 253-280.

73. Shapiro, Linda G., and George C. Stockman. *Computer vision*. Vol. 3. New Jersey: Prentice Hall, 2001.